\documentclass[10pt,twocolumn,letterpaper]{article}

\usepackage{times}
\usepackage{epsfig}
\usepackage{graphicx}
\usepackage{amsmath}
\usepackage{amssymb}
\usepackage{booktabs}
\usepackage{cite}

\newcommand{\wrt}{{\it w.r.t. }}    % with respect to
\newcommand{\eg}{\emph{e.g.}, }     % for example
\newcommand{\ie}{\emph{i.e.}, }     % that is
\newcommand\etc{\emph{etc.}}

\begin{document}

%%%%%%%%% TITLE
\title{Padding Aware Neurons}

\author{Dario Garcia-Gasulla\\
Universitat Politécnica de Catalunya - Barcelona TECH (UPC) \\
Barcelona Supercomputing Center (BSC)\\
{\tt\small dario.garcia@bsc.es}
\and
Victor Gimenez-Abalos\\
Barcelona Supercomputing Center (BSC)\\
% Plaça d'Eusebi Güell, 1-3, 08034 Barcelona\\
{\tt\small victor.gimenez@bsc.es}
\and
Pablo Martin-Torres\\
Barcelona Supercomputing Center (BSC)\\
{\tt\small pablo.martin@bsc.es}
}

\maketitle

\begin{figure}[t]
\centering
\includegraphics[width=0.45\textwidth]{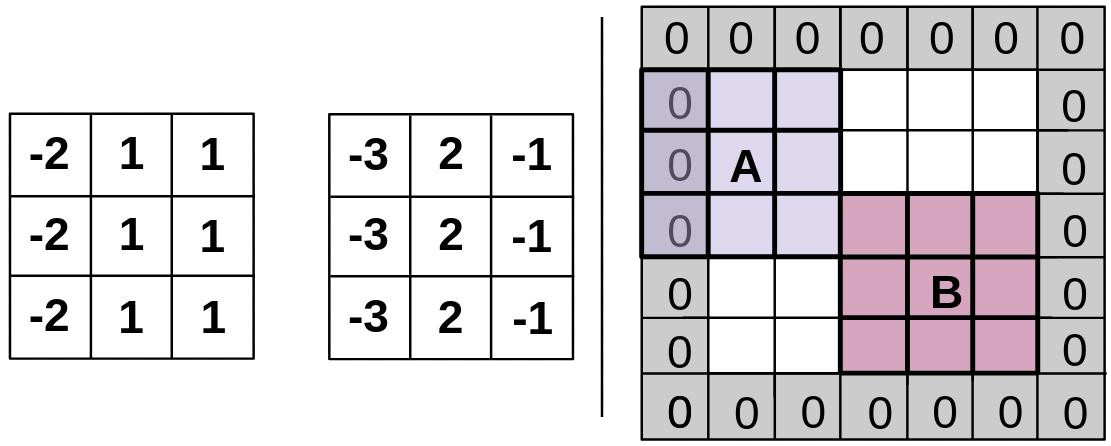} 
    \caption{On the left, example of two \textit{left} Padding Aware Neuron (PAN) filters. Activations on left-border locations (A) give larger outputs than in the centre (location B). On the right border, outputs are also slightly distinct. An actual neuron behaving analogously to the centre kernel can be appreciated in Figure~6.}
    \label{fig:source_pan}
\end{figure}

\begin{abstract}
Convolutional layers are a fundamental component of most image-related models. These layers often implement by default a static padding policy (\eg zero padding), to control the scale of the internal representations, and to allow kernel activations centered on the border regions. In this work we identify Padding Aware Neurons (PANs), a type of filter that is found in most (if not all) convolutional models trained with static padding. PANs focus on the characterization and recognition of input border location, introducing a spatial inductive bias into the model (\eg how close to the input's border a pattern typically is). We propose a method to identify PANs through their activations, and explore their presence in several popular pre-trained models, finding PANs on all models explored, from dozens to hundreds. We discuss and illustrate different types of PANs, their kernels and behaviour. To understand their relevance, we test their impact on model performance, and find padding and PANs to induce strong and characteristic biases in the data. Finally, we discuss whether or not PANs are desirable, as well as the potential side effects of their presence in the context of model performance, generalisation, efficiency and safety.
\end{abstract}

\section{Introduction}

Convolution has passed the test of time. Older than its competitors~\cite{fukushima1982neocognitron}, convolutional neurons have been successfully integrated with memory-based models (\eg LSTM~\cite{karim2017lstm}, GRU~\cite{zhang2018detecting}), attention-based architectures~\cite{yuan2021incorporating} and generative tasks~\cite{rombach2022high}. However, convolution has an undesired side-effect: the implicit reduction of internal representations~\cite{aghdasi1996reduction} caused by the impossibility of applying the convolved filter on border locations. To avoid this reduction, the most frequently used technique is \textit{padding}, adding synthetic data around the border of the input, so that kernels can activate there, and produce an output for every input.

The most popular padding type is, by far and wide, zero-padding (adding zeros to the input border). That is, a static padding, the same for every sample and location. Previous works noticed this constant signal adds a bias that reduces generalisation~\cite{mindthepad,nguyen2019distribution,aghdasi1996reduction,kayhan2020translation}, and several dynamic padding methods have been proposed to prevent it \cite{huang2021context,wu2019convolution,nguyen2019distribution,zhang2018detecting}, with very limited adoption~\footnote{https://pytorch.org/vision/stable/models.html\\https://www.tensorflow.org/resources/models-datasets}. The reason for this popularity is simple: models obtain better top-of-the-line metrics with static padding, when trained and tested on data from the same source. So far, the padding bias has been excused.

%In this work we study how static padding directly influences convolutional neurons, acting as a point of reference, a constant signal of the edge of data that convolutional neurons use to their advantage. As we will show, a significant amount of neurons adapt specifically to detect and convey this signal, passing on information about border location through the model, and influencing the majority of outputs. We name these neurons as \textit{padding aware neurons} (PANs). Our work shows how PANs are likely present in the vast majority of models trained with static padding, and proposes a methodology to locate them through the activations they produce. After analysing the particularities of its bias, we finalise with a brief discussion on the relevance of PANs for model generalisation, efficiency and safety.

In this work we dig deeper into how padding influences models. To do so, we provide evidence on how much model complexity is dedicated to the data edge bias (between 1\% and 3\%), and the magnitude of this shortcut in the model's outcome. This is characterized by the presence of \textit{padding aware neurons} (PANs), a symptom of padding bias. Our work shows how PANs are likely present in the vast majority of models trained with static padding, and proposes a diagnosis methodology which allows to locate them through their activation patterns.

\section{Setting}\label{sec:setting}
This work has been implemented using \texttt{PyTorch 1.12.0}~\cite{pytorch}, \texttt{torchvision 0.13.0}~\cite{torchvision}, \texttt{numpy 1.23.1}~\cite{numpy} and \texttt{scipy 1.8.1}~\cite{scipy}, the latter for Kolgomorov-Smirnov statistics. All models are provided pre-trained by \texttt{PyTorch}. These are:

\begin{itemize}
    \item ResNet-50~\cite{he2016deep}, trained on ILSVRC2012\cite{russakovsky2015imagenet}, named \textit{ResNet101\_Weights.IMAGENET1K\_V2} in {torchvision}.
    \item MobileNetV3~\cite{howard2019searching}, trained on ILSVRC2012, named \textit{MobileNet\_V3\_Large\_Weights.IMAGENET1K\_V2} in {torchvision}.
    \item GoogLeNet~\cite{szegedy2015going}, trained on ILSVRC2012, named \textit{GoogLeNet\_Weights.IMAGENET1K\_V1} in {torchvision}.
\end{itemize}

For each of these models, we analyse all convolutional layers with kernels bigger than 1x1. Notice these pre-trained models are frequently used as source for fine-tuning other models. 

We use a random batch from Caltech101~\cite{fei2004learning} in \S\ref{sec:def_and_ana}, for generating activations. In \S\ref{sec:pan_performance} we use the validation split of ILSVRC2012 for assessing bias. The code necessary to reproduce the experiments of this work can be found in https://gitlab.com/paper14/padding-aware-neurons.

\section{Definition \& Analysis}\label{sec:def_and_ana}

\begin{figure}[t]
    \centering
    \includegraphics[width=0.32\textwidth]{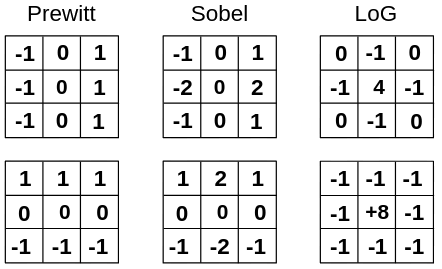} 
    \caption{Traditional edge detector filters. Prewitt (1st col.), Sobel (2nd col.) and Laplacian of Gaussian (3rd col.).}
    \label{fig:cv_filters}
\end{figure}

\textit{Padding aware neurons}, or PANs for short, are convolutional filters that learn to recognise the padding added to the input by some layers (\eg a convolutional layer). PANs pass information on border location through the network, introducing a spatial bias into the model which may or may not be desirable, depending on the domain of application~\cite{mindthepad}. Padding is often implemented as a vertical or horizontal edge (\eg zero padding), which makes PANs a type of edge detector. Edge detectors are fundamental vision kernels. The most popular ones include Prewitt, Sobel and the Laplacian of Gaussian (shown in Figure~\ref{fig:cv_filters}). These kernels look for value contrasts anywhere in the input\cite{maini2009study,7437397}, but are maximised when the value contrast is centred on the kernel (\eg centre square of a 3x3). This is visible in the symmetry exhibited by the filters of Figure~\ref{fig:cv_filters}. On the edges defined by padding, which are never centred on the kernel, edge detectors still activate moderately. In contrast to a regular edge detector, a PAN would maximize its output when the edge is located at the border of the filter, in order to discriminate the padding edges from other edges in the input. An example of one such kernels are shown in Figure~\ref{fig:source_pan}.

We hypothesise the existence of two types of PANs: nascent and downstream. Nascent PANs react when directly exposed to a padding area of the inputs of their layer, while downstream PANs react to the presence of padding as conveyed by PANs in previous layers (\ie they do not directly perceive padded values). In this work we focus on nascent PANs, which may have a configuration analogous to the kernel shown in Figure~\ref{fig:source_pan}.  %TODO DARIO REVISE
%The filter values of this PAN can be split in two. One part holds information (the signal or $s$), while the other learns to complement $s$ (the inhibitor or $i$). The padded input to the left illustrates two distinct positions (A \& B) in a 5x5 input with zero padding. Notice the kernel in Figure~\ref{fig:source_pan} activates highly when the padding is at the edge of the receptive field (\eg position A, where $i$ is cancelled by the zero padding), and lowly on any other central location (\eg position B, where $i$ cancels $s$). A specially low activation is produced when positioned on the opposite padding (\eg one column to the right of position B for the example of Figure~\ref{fig:source_pan}), as half of the signal is cancelled out by the padding. 
Beyond these toy examples, we consider any neuron that activates distinctively -- be it strongly or weakly -- on padded areas as a PAN. Notice a PAN can react to one \textit{or more} borders of the input. These include top row (T), bottom row (B), left-most column (L) and right-most column (R), but also any combination of these (\ie \texttt{T}, \texttt{B}, \texttt{L}, \texttt{R}, \texttt{TB}, \texttt{TL}, \texttt{TR}, \texttt{BL}, \texttt{BR}, \texttt{LR}, \texttt{TBL}, \texttt{TBR}, \texttt{BLR} and \texttt{TBLR}) in their non-overlapping definition (\eg \texttt{T} $\cap$ \texttt{BT} = $\emptyset$).

\subsection{Finding Edge Detectors}\label{sec:character}

Considering the complexities of characterising PANs through their high dimensional kernels~\cite{garriga2019,badenas2022}, we decide to use their activations instead. Next, we propose a method to identify nascent PANs by looking at the activations they produce on a padded input sampling. To be precise, we consider four padding regions of the input ($top$ and $bottom$ rows, $left$ and $right$ columns, all with corner overlap) of size one pixel on the short axis\footnote{Only the first/last row/column of the input guarantees the receptive field of the kernel covers the entire padded area, regardless of kernel size.}, and the remaining of the input ($centre$, with no overlap). We record the activations a given neuron produces on those five regions while processing a batch of in-distribution data. 

From these activations, we obtain five empirical probability density functions (PDF) per neuron ($A_{top}$, $A_{bottom}$, $A_{left}$, $A_{right}$, $A_{centre}$). By comparing every border PDF against $A_{centre}$ we obtain four Kolgomorov-Smirnov test (KS), which measure how distinct padding activations are for a given neuron. At this point its important to notice the sample size difference between border and center activations. $A_{top}$, $A_{bottom}$, $A_{left}$, $A_{right}$ all include the same number of values, $N$. $A_{centre}$ on the other hand includes $(N-2)^{2}$ activations, which grow quadratically \wrt $N$ assuming a stride of one.% Such that, $\forall N \geq 7 : (N-4)^{2} > N$, which is the case for most convolutional layers. The sample of KS value distributions is shown in Figure~\ref{fig:ks_dist}. In all layers, the majority of neurons have 

\begin{figure}[t]
    \centering
    \includegraphics[width=0.95\linewidth]{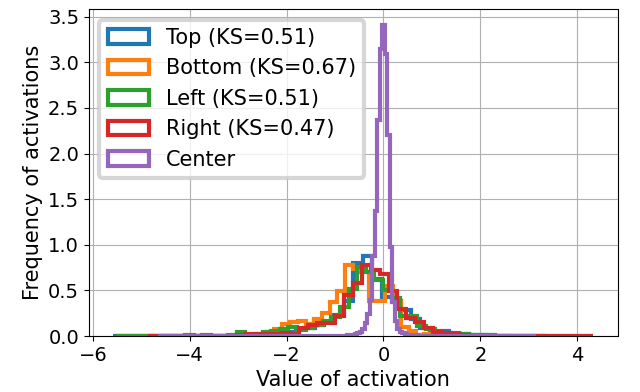}
    \includegraphics[width=0.95\linewidth]{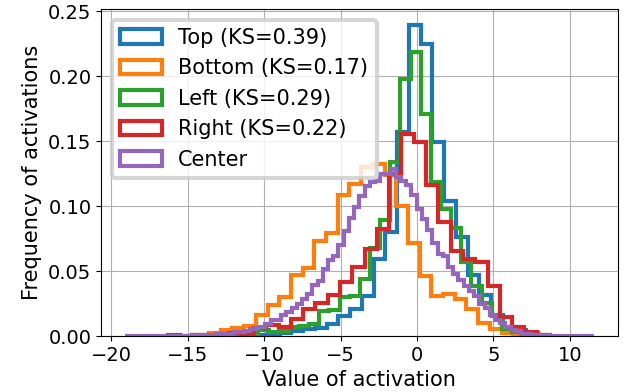}
    \caption{$A_{top}$, $A_{bottom}$, $A_{left}$, $A_{right}$ and $A_{centre}$ PDFs for two convolutional neurons of the ResNet50. Legend shows KS value of centre against every border region. Top plot: Neuron 51 from layer \textit{conv\_1}, an edge detector. Bottom plot: Neuron 101 from layer conv2\_2, a regular neuron.}
    \label{fig:ks_example}
\end{figure}

There is another difference between border and central activations. While border regions are entirely composed by edge data (the one defined by padding), central areas are partly so. While $A_{top}$, $A_{bottom}$, $A_{left}$ and $A_{right}$ contain only edge activations, $A_{centre}$ contains a majority of non-edge activations and a few data-driven edge activations. This skews the centre PDF \wrt the border ones, and turns the KS statistic into a measure of how distinctively are edge activations. A sort of \textit{padding-like edge detector}. Notice this method can not find edge detectors which are not straight vertical or horizontal. Figure~\ref{fig:ks_example} shows an example of border and centre PDFs for two neurons, together with the corresponding $KS$ values while using the two-sided KS, where the null hypothesis is that the two distributions are identical. 

\begin{figure}[t]
    \centering
    \includegraphics[width=0.99\linewidth]{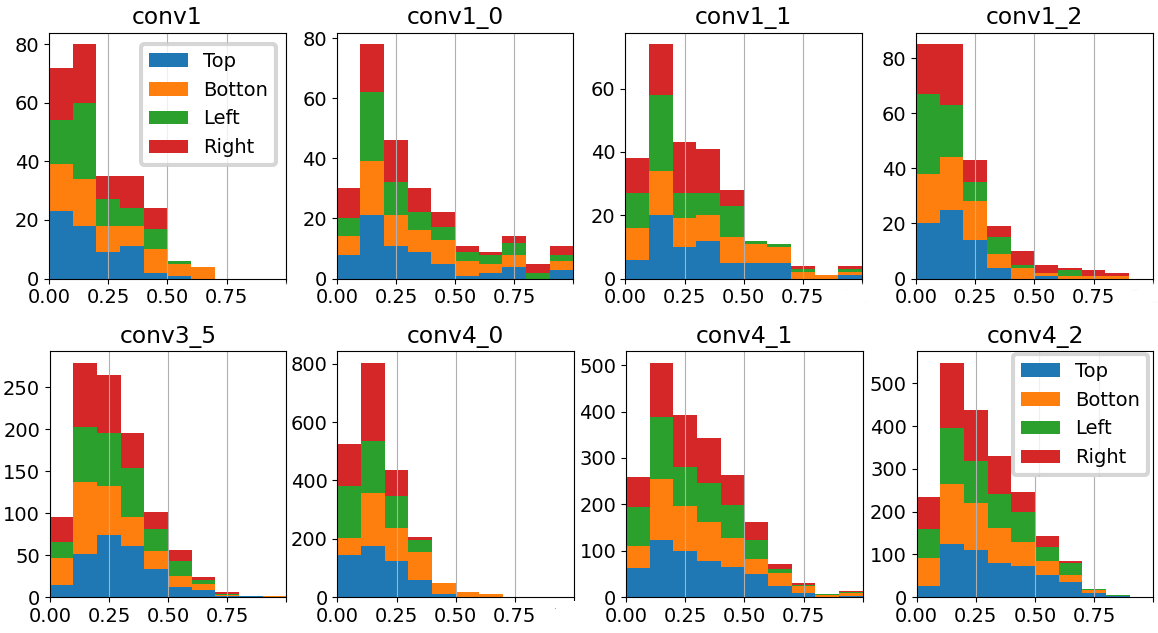}
    \caption{Stacked distribution of KS distances for the first and last four convolutional 3x3 layers of the ResNet50. Notice each neuron contributes with four values to each plot, $KS(A_{top},A_{centre})$, $KS(A_{bottom},A_{centre})$, $KS(A_{left},A_{centre})$ and $KS(A_{right},A_{centre})$.}
    \label{fig:ks_dist}
\end{figure}

Computing the KS values for all neurons in a model shows the overall activation divergence between centre and border locations. The KS distributions shown in Figure~\ref{fig:ks_dist} indicate most neurons have low KS values regardless of layer depth, with a mean KS between 0.1 and 0.3 on all cases. In other words, most convolutional neurons have no discriminative power between activations in a padded border and the centre. Notice each neuron contributes with 4 values to each plot of Figure~\ref{fig:ks_dist} ($KS(A_{top},A_{centre})$, $KS(A_{bottom},A_{centre})$, $KS(A_{left},A_{centre})$ and $KS(A_{right},A_{centre})$), which causes more KS values to be close to zero (\eg a vertical edge detector will most often generate low KS values for the top and bottom PDFs). Overall, results that indicate potential edge detector and PAN neurons (those with high KS values) are a minority found in most layers, regardless of depth.

\subsection{Finding PANs}

A KS test between the complete $A_{centre}$ and a border PDF cannot properly discriminate between PANs and the rest of edge detectors, as the presence of non-edge activations in $A_{centre}$ dominates its PDF. To discriminate PANs from regular edge detectors using the KS test, we need a distribution of $A_{centre}$ PDF which is comparable to border PDFs, that is, one which contains only edge activations. To that end, we define a simple hypothesis: the centre region of an input (of size $(N-1)^{2}$) will include \textit{at least} as many edges as a padded border (of size $N$). Notice this hypothesis, as well as the PDF reliability, grows weaker with the reduced input sizes typical of deeper layers.

%Leveraging this hypothesis we define an heuristic: we truncate $A_{centre}$ by keeping only the $k$ highest or lowest values of $A_{centre}$, where $k$ is the number of values in a padded border. We keep both the highest and lowest, since no assumptions are made on the relevance of their magnitude and sign for padding detection (\ie a PAN may detect padding by activating particularly strongly or weakly on it). For each end of the $A_{centre}$ distribution (which we refer to as $A_{centre}^{+}$ and $A_{centre}^{-}$), we use a different KS null hypothesis. For the most positive end of the centre, $A_{centre}^{+}$, we use the \textit{less} hypothesis ($KS^{+}$), and for the most negative end, $A_{centre}^{-}$, we use the \textit{greater} hypothesis ($KS^{-}$).

Leveraging this hypothesis we define an heuristic: we truncate $A_{centre}$ by keeping only the $k$ highest ($A_{centre}^{+}$) and $k$ lowest ($A_{centre}^{-}$) values of $A_{centre}$, where $k$ is the number of values in a padded border. We keep both the highest and lowest, since a PAN may detect padding by activating particularly strongly or weakly on it. For $A_{centre}^{+}$ we use the KS-test with the \textit{less} hypothesis ($KS^{+}$), \ie: $A_{centre}^{+}$ distribution is less than that of a margin (top, down, left or right), and for $A_{centre}^{-}$, we use the \textit{greater} hypothesis ($KS^{-}$), \ie as before but comparing with $A_{centre}^{-}$ instead. % TODO:DARIO REVIEW THIS

\setlength{\tabcolsep}{4pt}
\begin{table*}
\centering
\small{
\begin{tabular}{lcccccccccccccccccccc|c}
\textit{Model\textbackslash Depth}  & \textit{1}  & \textit{2}  & \textit{3} & \textit{4}  & \textit{5} & \textit{6} & \textit{7} & \textit{8} & \textit{9} & \textit{10} & \textit{11} & \textit{12} & \textit{13} & \textit{14}  & \textit{15} & \textit{16} & \textit{17}& \textit{18}& \textit{19}& \textit{20}& All\\
\hline
ResNet & 0 & 4 & 3 & 0 & 0  & \textbf{10}  & 1  & 3  & 0  & \textbf{16}  & 1   & 1   & 1   & 1    & 2    & \textbf{30}  &8  &- &- &- &81\\
                        & 0\% & \textbf{6\%} & 4\% & 0\% & 0\%  &\textbf{7\%}  & 0\%  & 2\%  & 0\%  & \textbf{6\%}  & 0\%   & 0\%   & 0\%   & 0\%  & 0\%  & 5\%    & 1\%  &- &- &- & 2.0\%\\
\hline
MobileNet & 0& 5& 1& 5& 0& 3& 2& 1& 5& 6& 3& \textbf{11}& 6& 9& \textbf{46}& \textbf{35}  &-  &- &- &-&138\\
                        &0\% & \textbf{31\%} & 1\% & \textbf{6\%} & 0\% & 2\% & 1\% & 0\% & 2\% & 3\% & 1\% & 2\% & 0\% & 1\% & \textbf{4\%} & 3\% &-  &- &- & - & 2.7\%\\
\hline
GoogLeNet & 0 & \textbf{8} & 2 & 1 & 0  & 0  & 0  & \textbf{8}  & 1  & 7  & 0   & 2   & 1   & 1    & 0    & \textbf{12}  & \textbf{10} & 5 & 0 & 0&58\\
                        & 0\% & 4\% & 1\% & 3\% & 0\%  &0\%  & 0\%  & \textbf{16\%}  & 0\%  & \textbf{10\%}  & 0\%   & 3\%   & 0\%   & 1\%    & 0\%    & \textbf{9\%} &3\%  &3\% &0\% &0\%&1.7\%\\
\end{tabular}}

\caption{Number of PANs found in different models, layer-wise. First row is absolute number of PANs, second row is percentage of PANs relative to layer size (rounded down). In bold, top three values per model on either category. Only 2D convolutional layers with kernels 3x3 or larger considered. Computed using $\theta=0.5$}
\label{tab:pan_distrib}
\end{table*}

\begin{figure}
  \centering
    \includegraphics[width=0.9\linewidth]{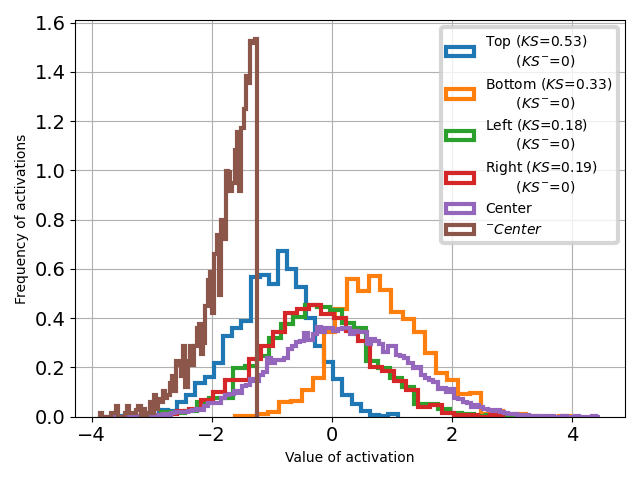}
    \caption{Histogram of neuron activations on the border regions, the center (purple) and the center truncated on the minus side (brown). Legend shows to Kolmogorov-Smirnov test. $KS$ corresponds to border vs center. $KS^{-}$ corresponds to border vs truncated center. Model: ResNet50. Layer: Conv3\_2. Neuron idx: 46.}
  \label{fig:truncated_example}
\end{figure}

The effect of using the truncated \textit{centre} PDF, is shown in Figure~\ref{fig:truncated_example}. The plot shows a neuron with negative activations for the top border, with the rest of activations being closer to zero. The computed $KS(A_{top},A_{centre})$ is 0.53. These results indicate this neuron is a vertical edge detector. However, when compared with the truncated $A_{centre}^{-}$, the same $A_{top}$ is no longer distinctive ($KS^{-}(A_{top},A_{centre}^{-})=0.0$), which indicates this neuron is not a PAN.

Given these insights, we label as PANs neurons which hold (1) a high $KS(A_{top|bottom|left|right},A_{centre})$ and, (2) a high $KS^{+}(A_{top|bottom|left|right},A_{centre}^{+})$ or a high $KS^{-}(A_{top|bottom|left|right},A_{centre}^{-})$. We set a threshold $\theta=0.5$ in the rest of the paper. $\theta$ can be modified to reduce or increase the requirements needed for PAN detection. The distributions of PANs identified using this methodology with $\theta=0.5$ is shown in Table~\ref{tab:pan_distrib}. % TODO DARIO REVISE: is the threshold picked due to some empirical decision or what? Affects relevance of 2% neurons found, with other thresholds it'd be different. Why 0.5?

On the models considered and with $\theta=0.5$, PANs represent roughly 2\% of all convolutional filters, and can be found at different depths. This may be caused by the explicit information about the presence of padding being lost or integrated (thus mixing with other activations) into other neurons after going through several layers. The disappearance of explicit padding information, however, does not preclude the information being used by the model, but it can motivate the model to periodically re-locate explicit padding so that the next few layers can more easily use that information. Later layers seem to include a remarkable amount of PANs, likely influenced by the large number of neurons found there. This could be influenced by the reduced reliability of the KS method when applied on inputs with small width and height, but it could also indicate padding location plays an important role on the final prediction.

Overall, applying the methodology to \textit{thousands} of filters yields \textit{hundreds} of edge detectors and \textit{dozens} of PANs per model. By slightly weakening the restrictions required to be labelled as a PAN their number can be easily doubled (\eg ResNet includes 193 PANs when using $\theta=0.4$).

%Figure~\ref{fig:ed_pan_distrib} shows how regular edge detectors and PANs synchronize their cycle, possibly to optimize their dual use (edge detectors provide information on padding, and PANs provide information on regular edges).

\subsection{PAN exploration}

Let us analyse neurons identified as PANs by the previously proposed method. For each neuron we look at their histogram of activations for the centre (complete and truncated PDF) and border regions. We also show these same plots, when inference is made replacing the zero padding policy by a reflect padding policy. Finally, we show activation maps for a couple of samples to understand its spatial response.

\begin{figure}
  \centering
    \includegraphics[width=0.98\linewidth]{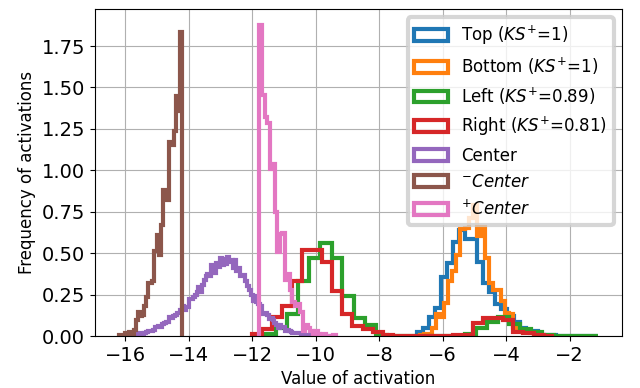}
    \includegraphics[width=0.98\linewidth]{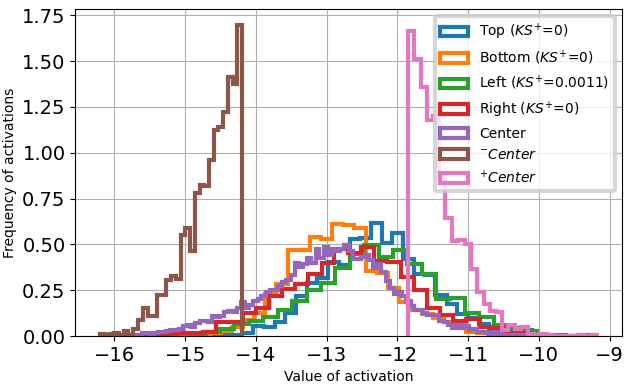}
    \includegraphics[width=0.25\linewidth]{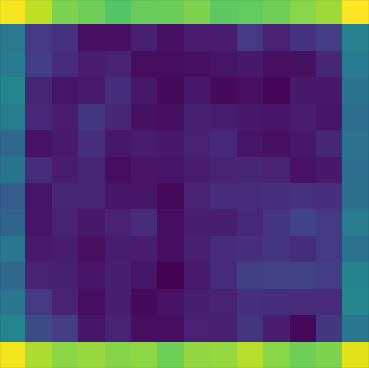}
    \hspace{15pt}
    \includegraphics[width=0.25\linewidth]{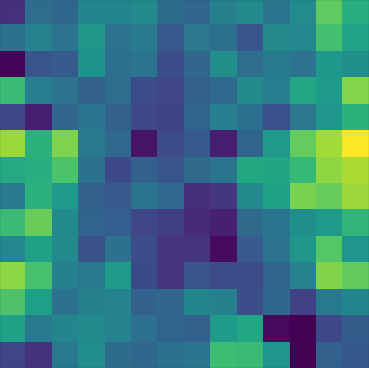}
    \caption{Top plot: Activation histogram of a PAN, where all four borders have high KS. Includes distributions for border regions, and central locations (complete and truncated). Legend shows KS confidence \wrt truncated distribution (\ie $KS^{+}(A_{border},A_{centre}^{+})$). Middle plot: Same as top, using padding reflect. Bottom plots: Activation heatmap on an input with zero and reflect padding. Model: ResNet50. Layer: Conv3\_1. Neuron idx: 41.}
  \label{fig:c31-41}
\end{figure}

The top plot of Figure~\ref{fig:c31-41} shows a PAN, with distinctively low activation values on all four borders, even when compared against the lowest values produced within the larger central area (\ie $A_{centre}^{+}$, in pink). With $\theta=0.5$, the PAN is detected as \texttt{TBLR}. An inspection of the activations produced by the kernel on two inputs (bottom plot of Figure~\ref{fig:c31-41}) shows how this PAN has a preference for the bottom and top padding, which is consistent with $KS^{+}(A_{left|right},A_{centre}^{+}) < KS^{+}(A_{top|bottom},A_{centre}^{+})$ (as shown in the top plot). Notice $A_{left}$ and $A_{right}$ have a bimodal distribution, peaking both at -10 and at -4. This is caused by particularly strong activations on corner positions, which are high even within $A_{top}$ and $A_{bottom}$. This neuron, beyond being padding aware, is also corner aware, a behavior found on other neurons (\eg conv1\_0, 17; conv3\_1, 212; conv4\_1, 296; conv4\_2, 447). When the padding is changed from \textit{zero} to \textit{reflect}, as shown in the middle plot of Figure~\ref{fig:c31-41}, the neuron no longer detects padding. The distributions of activation values for border regions become indistinguishable from the distribution in the centre.

\begin{figure}
  \centering
    \includegraphics[width=0.99\linewidth]{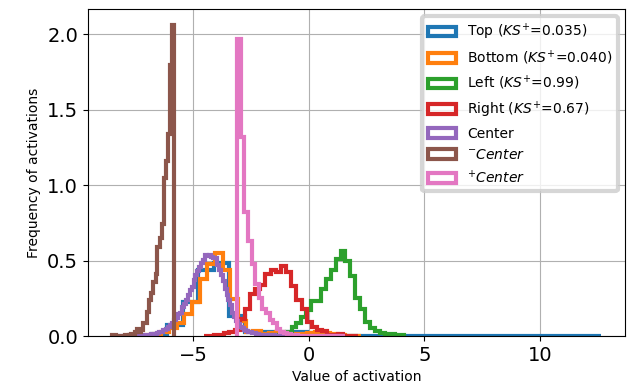}
    \includegraphics[width=0.99\linewidth]{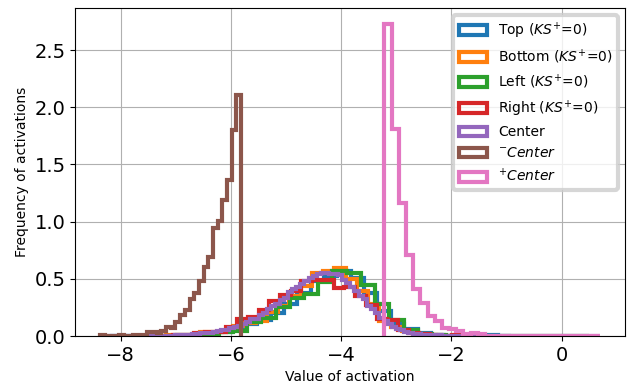}
    \includegraphics[width=0.25\linewidth]{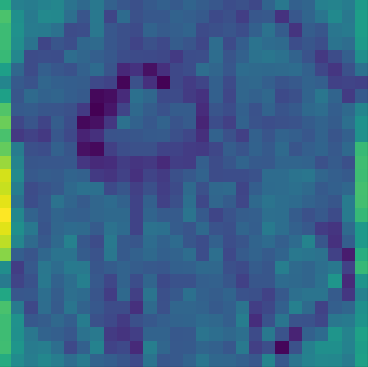}
    \hspace{15pt}
    \includegraphics[width=0.25\linewidth]{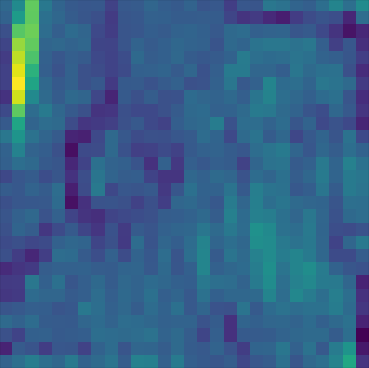}
    \caption{Top plot: Activation histogram of a PAN, where the left and right borders have high KS. Includes distributions for border regions, and central locations (complete and truncated). Legend shows KS confidence \wrt truncated distribution (\ie $KS^{+}(A_{border},A_{centre}^{+})$). Middle plot: Same as top, using padding reflect. Bottom plots: Activation heatmap on an input with zero and reflect padding. Model: ResNet50. Layer: Conv2\_1. Neuron idx: 67}
  \label{fig:c21-67}
\end{figure}

Another representative neuron is shown in Figure~\ref{fig:c21-67}. In this case the PAN activates distinctively high on the left and right padding. Since $A_{left}$ is significantly higher than $A_{right}$, this may be primarily a \texttt{L} PAN that also detects the right border by complement. This is in fact a behaviour compatible with the kernel shown at the centre of Figure~\ref{fig:source_pan}. For the top and bottom padding locations, this neuron's activations are indistinguishable from those on central locations. The long tail of the top and bottom distributions speaks of potential corner detection capabilities. All this is illustrated by the bottom plot of Figure~\ref{fig:c21-67}, which shows activations on two inputs. Notice some edges are detected in centre locations, but not as strongly as on the left and right padding. The middle plot of Figure~\ref{fig:c21-67} shows the same activations when zero padding is replaced by reflect padding. When this is the case, the neuron no longer detects padding, with $A_{left}$ and $A_{right}$ becoming aligned with the rest of distributions.

\begin{figure}
  \centering
    \includegraphics[width=0.99\linewidth]{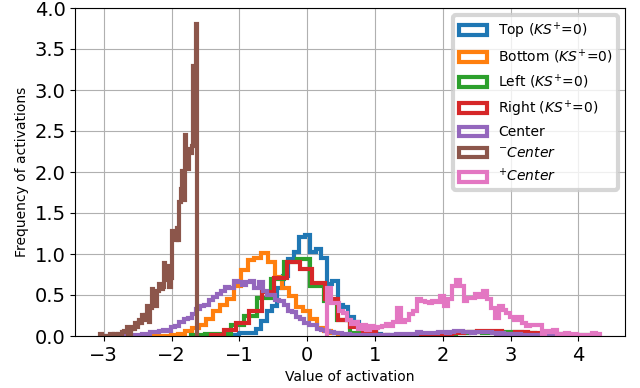}
    \includegraphics[width=0.99\linewidth]{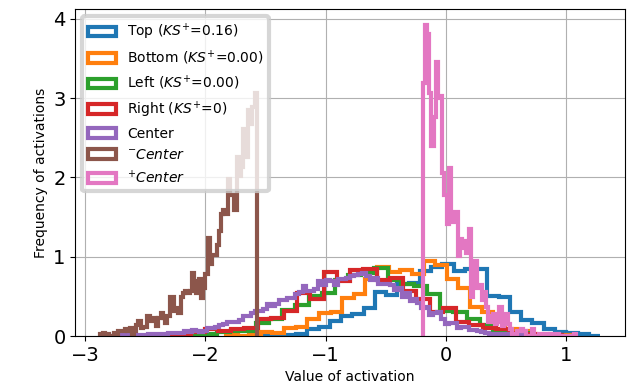}
    \includegraphics[width=0.25\linewidth]{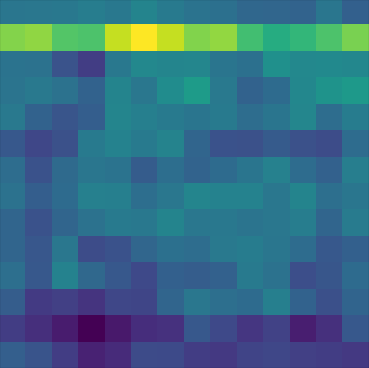}
    \hspace{15pt}
    \includegraphics[width=0.25\linewidth]{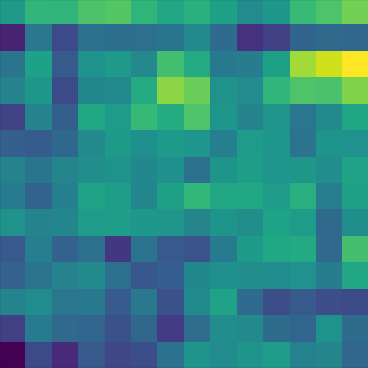}
    \caption{Top plot: Activation histogram of a neuron which is an edge detector candidate and a potential downstream PAN, not detected as PAN. Includes distributions for border regions, and central locations (complete and truncated). Legend shows KS confidence \wrt truncated distribution (\ie $KS^{+}(A_{border},A_{centre}^{+})$). Notice the truncated centre distribution on the high side (pink) is bimodal, with one peak around \textit{zero} and one around \textit{two}. Middle plot: Same as top, using padding reflect instead of zero. Bimodal distribution disappears. Bottom plots: Activation heatmap on an input with zero and reflect padding. Model: ResNet50. Layer: Conv3\_2. Neuron idx: 158}
  \label{fig:c32-158}
\end{figure}

The last neuron discussed here is the \textit{downstream} PAN of Figure~\ref{fig:c32-158}. Following the proposed methodology, this neuron is detected as a potential edge detector ($KS(A_{top},A_{centre})=0.66$), but not as a PAN ($KS^{+}(A_{top},A_{centre})=0.0$) (see top plot). Its spatial activations on two different inputs (bottom plots of Figure~\ref{fig:c32-158}) indicate this is no regular edge detector. It activates distinctively on the \textit{second} highest row of the input, as if it was detecting the top padding from afar. This explains the bimodal behaviour of this neuron in the top plot, where the truncated $^{+}centre$ distribution (which includes most of the second row) peaks both at around two (activations of the second highest row) and zero (activations on the rest of centre). Since the kernel of this neuron is 3x3, it cannot directly detect the padding from this location (\ie on the second highest row activations, the kernel is located entirely on the unpadded input). This neuron gets the information about image border location from a previous layer, and turns off (see middle plot of Figure~\ref{fig:c32-158}) when static padding is removed. 

\subsection{Nascent PAN types}

Through the analysis defined in the previous sections we have characterised and identified several types of nascent PANs, those that directly detect padding in the input. Nascent PANs frequently have a multi-modal behaviour, detecting two or more padding edges. This multi-border detection can be generic (\ie several borders detected indistinguishably), or it can be distinct for different border types. The neuron shown in Figure~\ref{fig:c31-41}, for example, can discriminate between horizontal borders (top and bottom), vertical borders (left and right) and the rest of the input. But it cannot discriminate among horizontal borders (between top and bottom padding), or among vertical ones (left and right padding). On the other hand, the neuron shown in Figure~\ref{fig:c21-67} can discriminate between left and right padding. This later behaviour is consequence of the asymmetrical kernels PANs may have, exemplified in the kernels of Figure~\ref{fig:source_pan}.

\begin{table*}
\centering
\small{
\begin{tabular}{lccccccccccccccc}
PAN type         & \texttt{T}  & \texttt{B}  & \texttt{L} & \texttt{R}  & \texttt{TB} & \texttt{TL} & \texttt{TR} & \texttt{BL} & \texttt{BR} & \texttt{LR} & \texttt{TBL} & \texttt{TBR} & \texttt{TLR} & \texttt{BLR}  & \texttt{TBLR} \\
\hline
ResNet & 10 & 32 & 8 & 10 & 8  & 1  & 0  & 0  & 3  & 4  & 1   & 0   & 0   & 0    & 4     \\
MobileNet & 47 & 90 & 9 & 6 & 9  & 5  & 0  & 1  & 3  & 8  & 0   & 0   & 1   & 1    & 13      \\
GoogLeNet & 7 & 24 & 4 & 2 & 7  & 1  & 0  & 0  & 1  & 7  & 0   &0    &  1  &    1 & 3   \\
%ResNet50 ($0.4$) & 20 & 65 & 9 & 20 & 9  & 6  & 0  & 1  & 2  & 6  & 0   & 0   & 1   & 0    & 6    &   \\
\end{tabular}}
\caption{Distribution of PAN types identified on different models, with $\theta=0.5$.}
\label{tab:pan_subtypes}
\end{table*}

We identify 14 possible types of nascent PANs based on which padding borders they detect (\ie \texttt{T}, \texttt{B}, \texttt{L}, \texttt{R}, \texttt{TB}, \texttt{TL}, \texttt{TR}, \texttt{BL}, \texttt{BR}, \texttt{LR}, \texttt{TBL}, \texttt{TBR}, \texttt{BLR} and \texttt{TBLR}). We study the distribution of nascent PAN types with the proposed method in Table~\ref{tab:pan_subtypes}. Single border detectors (\ie \texttt{T}, \texttt{B}, \texttt{L}, \texttt{R}) are the most frequent types, representing about 75\% of all identified PANs. The rest are mostly PANs which can detect complementary borders (\ie \texttt{TB}, \texttt{LR}), or all four borders (\ie \texttt{TBLR}). Complementary borders detecting PANs are likely to be mirrored variations of the kernel shown in the middle of Figure~\ref{fig:source_pan}, while the four borders PAN may be asymmetrical versions of the bottom Laplacian of Gaussian filter shown in Figure~\ref{fig:cv_filters}.

\section{Performance and Bias}\label{sec:pan_performance}
Once we have established the existence and pervasiveness of PANs in models trained with zero padding, let us now assess the role these neurons play in model behaviour. To do so, we study their influence in the network output using four versions of the same pre-trained ResNet50, without fine-tuning:

\begin{itemize}
    \item The \textit{original} model, using the default zero-padding.
    \item The \textit{reflect} model, where the padding of all convolutional neurons has been changed to PyTorch's reflect.
    \item The \textit{PAN-reflect} model, where the padding of the neurons identified as PANs by the previous methodology (for ResNet50, 2.0\% of convolutional neurons, 81 overall) has been changed to reflect. The rest of neurons preserve zero-padding.
    \item The \textit{RAND-reflect} model, where the padding of randomly sampled non-PANs has been changed to reflect and the rest preserve zero-padding. The random subset has the same size (2.0\% of neurons) and follows the same layer distribution as \textit{PAN-reflect}. This is the control set.
\end{itemize}

We use the quantitative differences in the outputs of these models to study the impact padding has towards specific classes (\ie the amount of padding bias). Then, we study the influence of PANs in the context of particular data samples.

\subsection{Bias influence}\label{sec:biasinfluence}
To verify to which extend PANs add relative location bias to the model, we compare the soft-max outputs of \textit{original} with those of \textit{PAN-reflect}. To be precise, we compute the odds of the prediction probability for each class. Assuming samples to be \texttt{i.i.d.}, this can be computed as the quotient of the sum of soft-max outputs for all images $i$ in the dataset:

\begin{equation*}
    Odds(c) = \frac{P(c|M_{Pan-reflect})}{P(c|M_{original})} = \frac{\sum_i M_{Pan-reflect}(i)[c]}{\sum_i M_{original}(i)[c]}
\end{equation*}

And analogously for \textit{RAND-reflect}. For \textit{PAN-reflect}, odds above $1$ for a class $c$ indicate a higher confidence in the prediction of $c$ in the absence of PANs. This can also be interpreted as padding being used as evidence against that class. Conversely, values below $1$ would imply padding is being used as evidence toward the class. 

\begin{figure}
    \centering
    \includegraphics[width=0.9\linewidth]{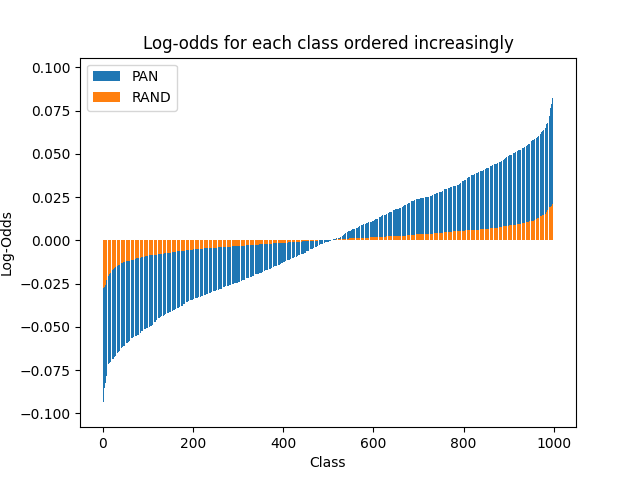}
    \caption{Ordered change in log-odds, for \textit{PAN-reflect} and \textit{RAND-reflect} \wrt original model. Vertical axis is the amount of change. $\pm 0.05$ log-odds corresponds to 5\% difference in odds.}
    \label{fig:lodds}
\end{figure}

Figure~\ref{fig:lodds} presents the logarithm of the odds per class, computed on the ILSVRC validation set for both \textit{PAN-reflect} and \textit{RAND-reflect}. All classes are affected, a few severely so. Table~\ref{tab:odds} lists all classes whose odds change by more than 7\%. We choose a threshold instead of the top-K to illustrate how the odds change in an asymmetrical manner: there are more classes which use padding as evidence toward the class (odds $<$ 1) than those that use it against. Remarkably, classes for which padding is used as evidence against it seem to be mostly fine-grained types (mainly animal species and dogs, with the exception of \textit{sliding door}), which hints at the relevance of padding for overfitting. Conversely, there are no animals among the classes that use padding as positive evidence. Using a 5\% threshold yields consistent results: out of the 111 classes with negative log odds, the only animal is the \textit{English Foxhound}, whereas for the 99 classes with positive log odds, there are only five classes which are not fine-grained animals.
%the only exceptions are, in decreasing order of odds: \textit{sliding door}, \textit{website}, \textit{home theater}, \textit{safe}, and \textit{china cabinet}.

\begin{figure}
    \centering
    \includegraphics[width=0.9\linewidth]{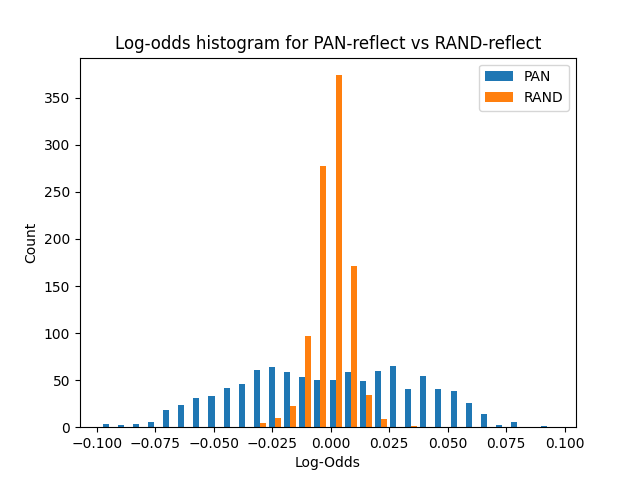}
    \caption{Class histogram of log odds change \wrt original model, computed for both \textit{PAN-reflect} and \textit{RAND-reflect}. Notice how the former has both a wider range and a bimodal distribution.}
    \label{fig:logoddshist}
\end{figure}

To verify if findings are related with the relevance of padding or with the noise added by the data distribution, let us consider the results while using \textit{RAND-reflect} (orange in Figures~\ref{fig:lodds} and \ref{fig:logoddshist}). In this case, the distribution of PANs' odds is characteristically different from that of random, similarly-sampled neuron sets. While PANs seem to affect most classes to a large degree, either positively or negatively, the random set effect on classes is very limited. Only a few classes are affected, with the most common result being no output change. These results indicate PANs strongly and homogeneously alter most classes' prior, whereas an equally sized random subset of neurons does not. 

Repeating this experiment with model \textit{reflect} changes the input distribution of 100\% of convolutional layers, whereas the previous two experiments (with \textit{PAN-reflect} and \textit{RAND-reflect}) changed only 2\% of neurons. As a result, the \textit{reflect} odds suffer more extreme changes than either one of the above. No tendency around which classes receive positive and which negative log odds was found. In this particular experiment, we believe the larger odds variance has to do with noise added to the distributions, rather than due to some intrinsic quality of how padding is used.

\begin{table}[]
    \centering
    \small{
    \begin{tabular}{c|c|c|c}
        Class & Odds & Class & Odds \\ 
         \hline
drum & 0.91 & cheetah & 1.10 \\ 
muzzle & 0.91 & Norfolk Terrier & 1.09 \\ 
packet & 0.91 & sliding door & 1.09 \\ 
sunscreen & 0.92 & Irish Water Spaniel & 1.08 \\ 
barrette & 0.92 & box turtle & 1.08 \\ 
tandem bicycle & 0.92 & Dobermann & 1.08 \\ 
candle & 0.92 & Flat-Coated Retriever & 1.08 \\ 
tent & 0.92 & Alaskan Malamute & 1.08 \\ 
tray & 0.92 & gossamer-winged butterfly & 1.07 \\ 
comic book & 0.93 & West Highland White Terrier & 1.07 \\ 
Windsor tie & 0.93 & Greater Swiss Mountain Dog & 1.07 \\ 
tile roof & 0.93 & guenon & 1.07 \\ 
backpack & 0.93 &&\\ 
overskirt & 0.93 &&\\ 
buckle & 0.93 &&\\ 
lab coat & 0.93&& \\ 
shoal & 0.93 &&\\ 
paper knife & 0.93&& \\ 
whistle & 0.93 &&\\ 
ice pop & 0.93 &&\\ 
stethoscope & 0.93 &&\\ 
barbell & 0.93 &&\\ 
lakeshore & 0.93 &&\\ 
megalith & 0.93 &&\\ 
scarf & 0.93 &&\\ 
    \end{tabular}}
    \caption{Classes with odds beyond 7\% computed between the \textit{PAN-reflect} and the \textit{original} model. Classes with odds above one, increase their confidence in the absence of PAN information, are less frequent and are mostly composed of animals. For odds below 1, we check $\frac{1}{odds(c)}>1.07$.}
    \label{tab:odds}
\end{table}

\subsection{Sample influence}
The previous section shows a clear influence of padding in the overall performance and behaviour of the model. However, the class-scale at which analysis is made means that the effect of PANs on single predictions is aggregated in the mean for each class. To analyse this facet, we look for the individual samples with the largest change in the network's output. We compute this change as the Manhattan distance between the logits of the \textit{original} and the \textit{PAN-reflect} model.

Significantly, the 30 images with the biggest padding influence are all incorrectly classified by both the \textit{original} and the \textit{PAN-reflect} models, the predicted class remaining the same. Analysing the top 5,000 most affected images (10\% of the whole dataset), we find that the number of disagreements between models is remarkably low (67 images). The limited impact PANs have on samples which are not part of the model training set, could also be the result of padding information being used for overfitting particularly hard training samples. 

When repeating the experiment on \textit{RAND-reflect}, these effects disappear. The sample with the 5000th highest divergence with the \textit{PAN-reflect} has around 4 distance units, whereas for \textit{RAND-reflect} with this distance happens on the 13th sample. This alone shows \textit{PAN-reflect} affects with more strength to orders of magnitude more samples than \textit{RAND-reflect}. Of those 13 samples, 12 of them are incorrectly predicted as \textit{tench}, which indicates the preference of these randomly chosen 2\% of neurons for this class.% despite not being particularly challenging, and the ground truth are several other classes. 

% Currently done experiments: validation acc of a network trained with pad0, using pad0 vs using pad-reflect and others; show accuracy changes (Acc difference boxplot or sth?)
% Then, repeat experiment using pad0, but intercepting the individual outputs of the 25 most likely PANs by changing the values in the output. (TODO: maybe for a gaussian or sth? or a padding reflect or whatever?) and analysing the effect of changing these few neurons wrt to changing any other 25 neurons in the same manner. 
% Correlate class acc decay with a scatterplot to show these neurons are the guilty of the accuracy decay, illustrating the point. (scatterplot is horrible, need to check other possible plots)

% Justification: One of the things we need to show is that PANs matter for output. If what we say is true, PANs are the set of neurons that require padding for making the system work as intended. We show in experiment X that there's noticeable accuracy changes when switching the padding type. We now want to show that the change is focused in the PAN neurons. To do so, we inhibit XX neurons (identified as PAN through previous processes) by changing their output on the padding output.

\section{Discussion}

The use of static padding in convolutional layers provides the model with a stable signal of a perceptual edge. That much was known from previous works~\cite{mindthepad,nguyen2019distribution,aghdasi1996reduction,kayhan2020translation}. This paper reveals the extent of this inductive spatial bias, identifying a set of neurons specialized in locating and exploiting it (what we call PANs), which account for at least 1.5\%-3\% of all deep CNNs convolutional filters. Considering PANs are likely to be inheritable (as long as the fine-tuned model keeps zero padding) and the fact that PANs were found on popular pre-training sources, one can assume PANs are a widespread phenomenon.

Experiments indicate padding information is used to change the prior of most classes. PANs seem to be used as evidence against fine-grained classes (\ie animals), and seldom as evidence for them. For the ILSVRC task we derive two different hypothesis for explaining this. Either samples from fine-grained class are generally better framed, which keeps the padding away from the patterns most relevant for the class, resulting in a spatial bias that can be leveraged; or padding is used as a reference to identify arbitrary patterns in particularly hard samples, helping overfit on examples from the long tail~\cite{d2021tale}. Testing both these hypothesis remains future work as it requires its own experimental setup.

%Our current hypothesis is that padding influence is limited in the presence of other learnt patterns of the model. In the absence of patterns that help the model classify, padding is used as strong evidence in the model's prediction. This is further supported by the fact that cases in which the \textit{original} and \textit{PAN-reflect} models have significantly different logits, both models' predictions tend to be much different.

The desirability of PANs in a model depends on the application, and its definition of un/desirable bias. On tasks with fixed framing (\eg fundus retina images~\cite{zapata2020artificial}, static cctv feed~\cite{bhatti2021weapon} \etc) PANs may provide a useful location reference allowing a better contextualisation and structuring of the input. On tasks which entail frame freedom (\eg objects in the wild, variances among devices) PANs learn an arbitrary bias, which may contribute to overfitting and lack of generalisation~\cite{mindthepad,nguyen2019distribution,aghdasi1996reduction}. For these reasons, we recommend practitioners to choose padding carefully, using dynamic padding (such as reflect) by default but accounting for the removal of PAN information. It remains to be seen how the presence of PANs affect in downstream tasks different to classification, such as training or fine-tuning for object detection, or whether fine-tuning in a new task removes or adds more PANs.

Even when PANs are useful, their current design is not efficient. A lot of parameters (PAN kernels) \textit{and} computation are wasted on recognizing a constant. For those cases where PANs are indeed desirable, one may find more efficient and rich versions of them, at least in three different ways: (1) by implementing a sparse computation which skips padding products, (2) by using models with pre-initialised PAN kernels spread along the model, and (3) by adding the complementary axis information to padding (row height in vertical padding and vice-versa) for complete spatial reference.

Finally, let us consider a safety vulnerability PANs entail. Given their characteristic pattern, PANs are easy to fool and trigger. Adding a one-row/column of zeros anywhere in the input will cause PANs to fire, out of the manifold and into unpredictability. This can be easily mitigated, for example, by doing data augmentation during training with random rows/columns of padding \textit{in} the input. This is strongly suggested for models deployed on critical domains.

%ACKs: The authors would like to acknowledge the work of Carles Riera[] & Victor Badenas [] who found preliminary evidence on the existence of PANs through weight and gradients analysis respectively. 

\section{Acknowledgements}
We acknowledge previous collaborations with Carles Garriga and Victor Badenas, trying to locate PANs using kernel values. We would also like to acknowledge Ferran Parés for early discussions in the conceptualization of this work. This work has been partially funded by the project SGR-Cat 2021 HPAI (AGAUR grant n.01187). % PENDING TO CHECK IF IT'S THIS ONE OR KNOWLEDGE

%%%%%%%%% REFERENCES
{\small
\bibliographystyle{ieee_fullname}
\bibliography{references}
}
\end{document}